\documentclass[runningheads]{llncs}

\usepackage{accv}

\usepackage{accvabbrv}
\usepackage{float}
\usepackage{graphicx}
\usepackage{booktabs}
\usepackage{bbm}
\usepackage[accsupp]{axessibility}

\usepackage{hyperref}

\usepackage{orcidlink}

\begin{document}
\raggedbottom
\title{Quo Vadis, Driving Scene Reconstruction? Occlusion-Aware Gaussian Densification}

\titlerunning{Occlusion-Aware Gaussian Densification}

\author{Dung Do Tien\inst{1,2}\fnmsep\thanks{Equal contribution.}\orcidlink{0009-0001-6465-5941} \and
Kien Nguyen Do Trung\inst{1,2,\star}\orcidlink{0009-0004-0377-0059} \and
Kiet Dang Vu\inst{1,2}\orcidlink{0009-0004-0964-8134} \and
Trung Thai Tran\inst{1,2}\orcidlink{0009-0002-1422-9685} \and
Duc Dung Nguyen\inst{1,2}\fnmsep\thanks{Corresponding author: \href{mailto:nddung@hcmut.edu.vn}{nddung@hcmut.edu.vn}}\orcidlink{0000-0001-7321-7401}}

\authorrunning{Dung Do Tien et al.}

\institute{AITech Lab., Ho Chi Minh City University of Technology (HCMUT) \and
Vietnam National University Ho Chi Minh City (VNUHCM)}

\maketitle

\begin{abstract}
While recent feed-forward 3D Gaussian Splatting methods enable efficient driving-scene reconstruction from surround-view images, their reliance on a single observation fundamentally hinders novel-view synthesis. Regions occluded in the input can become visible from novel viewpoints, leaving the reconstructed Gaussians under-constrained and incomplete. Moving beyond image-space inpainting for rendered views, we introduce OccSplat, a feed-forward framework whose core contribution is Occlusion-Aware Gaussian Densification, explicitly targeting missing scene geometry rather than missing pixels. By predicting new Gaussian primitives for unsupported regions, OccSplat completes the scene directly in the 3D representation. Specifically, OccSplat first identifies occluded regions that lack Gaussian support through Gaussian Occlusion Mapping. Gaussian Occlusion Completion then generates new primitives for these regions. The predicted primitives are integrated into the original representation to improve novel-view synthesis while retaining feed-forward inference without per-scene optimization. Experiments on the nuScenes benchmark demonstrate that OccSplat improves on the compared methods and yields particularly large gains in occluded regions, leading to more accurate and visually coherent driving-scene reconstruction.

\end{abstract}

\section{Introduction}
\label{sec:intro}

Photorealistic reconstruction and novel-view synthesis of driving scenes from
vehicle-mounted cameras are foundational capabilities for autonomous driving,
underpinning closed-loop simulation, safe evaluation of perception and planning stacks,
and large-scale data augmentation for rare or safety-critical events. The recent
emergence of 3D Gaussian Splatting (3DGS)~\cite{3dgs}, which represents a scene with an
explicit set of anisotropic Gaussians and renders through fast differentiable
rasterization, enables fast, high-fidelity rendering after scene optimization and has
become a widely used representation for this setting.

Building on 3DGS, a growing body of work specializes the representation to driving
scenes~\cite{zhou2024drivinggaussian,yan2024street,huang2024textit,khan2025autosplat},
modeling the static background, dynamic actors, and unbounded street layout that
characterize on-road data. These methods achieve impressive rendering quality, but are
fundamentally \emph{per-scene optimization} approaches: each scene must be fit
individually, typically requiring densely overlapping views, accurate camera extrinsics,
and minutes to hours of optimization. The need to optimize each new scene limits throughput in large-scale
autonomous-driving pipelines.
\begin{figure*}[t!]
  \centering
  \includegraphics[width=\linewidth]{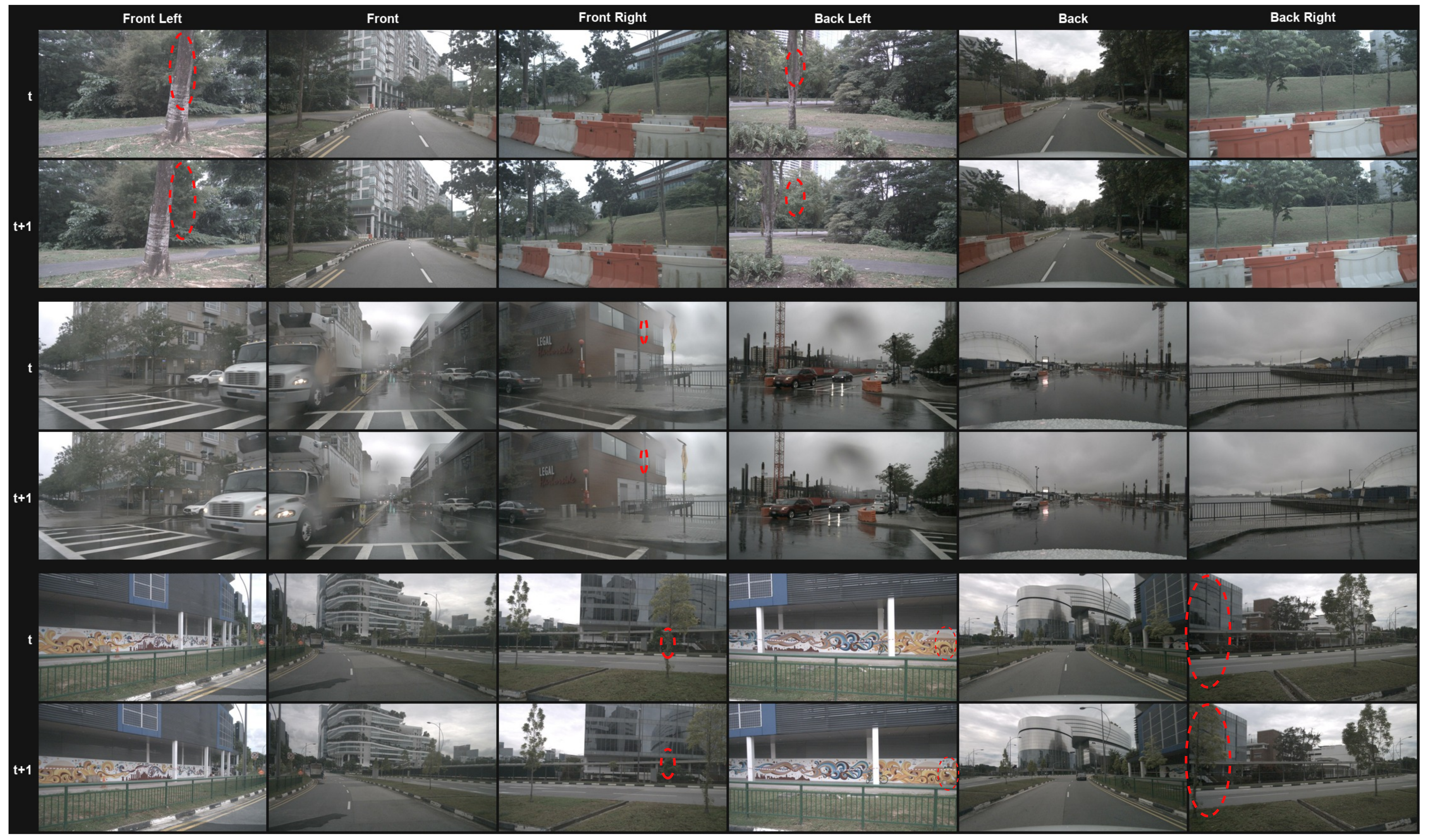}
  \caption{%
    \textbf{The disocclusion problem in forward-driving single-frame (SF) reconstruction.}
    Six surround-view cameras at a source frame $t$ (top) and target frame $t{+}1$
    (bottom). As the ego-vehicle advances, content hidden behind foreground structures at
    $t$ becomes \emph{disoccluded} at $t{+}1$ (red dashed ellipses). A model that only
    observes frame $t$ has no direct visual evidence for these regions.
  }
  \label{fig:teaser}
\end{figure*}

To lift these constraints, \emph{feed-forward} (generalizable) Gaussian Splatting
predicts pixel-aligned Gaussian primitives directly from images in a single forward pass, enabling training on large corpora and inference without per-scene optimization on unseen
scenes~\cite{charatan2024pixelsplat,chen2024mvsplat,chen2024mvsplat360,huang2025no,kang2025selfsplat,szymanowicz2024splatter,xu2025depthsplat,ye2025no,zhang2024gs}.
DrivingForward~\cite{tian2025drivingforward} adapts this paradigm to driving, jointly
learning pose, depth, and Gaussian networks in a self-supervised manner so that
surround-view scenes can be reconstructed without depth ground truth or camera extrinsics.
We focus on its challenging single-frame (SF) setting:
the model observes only the surround-view images of the current frame $t$ and must
synthesize the surround views of the future frame $t{+}1$.

As the ego-vehicle drives forward between $t$ and $t{+}1$, the camera rig translates along
the direction of travel and the effective viewpoint changes. Consequently, scene regions
that are \emph{occluded} at $t$, hidden behind trees, poles, fences, or building edges,
become \emph{disoccluded} at $t{+}1$ (Fig.~\ref{fig:teaser}). Feed-forward pixel-aligned
methods place a Gaussian only where there is an observed pixel to lift it from, so surfaces
with no visible support at $t$ receive no primitives at all. When these surfaces are
exposed at $t{+}1$, the renderer has nothing to draw in their place, yielding holes and
stretched ``rubber-sheet'' artifacts. This disocclusion gap is intrinsic to forward-driving SF synthesis: it
arises from viewpoint changes that reveal content absent from the source images.

\emph{Quo vadis}? Where, then, is the scene going? The question is literal as well as
rhetorical: the model must reason not only about what it currently sees, but about the
content that its forward motion is about to uncover. We therefore recast future-frame
driving reconstruction as an \textbf{Occlusion-Aware Gaussian Densification} problem and
instantiate it in \textbf{OccSplat}, a feed-forward surround-view framework that
anticipates and completes soon-to-be-revealed content before the ego-vehicle uncovers it.
Starting from a pixel-aligned Gaussian representation of the current frame, OccSplat first
localizes regions that lack Gaussian support and are therefore prone to disocclusion, then
densifies them with new 3D Gaussian primitives that plausibly complete the disoccluded
geometry and appearance. Because these primitives are fused into the same representation as the observed scene, the completed content can be rendered from a shared 3D representation instead of
being patched independently in each image. This
densification is a single additional feed-forward pass on top of the backbone, closing the
disocclusion gap while retaining feed-forward inference without per-scene optimization.

Our contributions are summarized as follows:
\begin{itemize}
  \item We introduce \textbf{OccSplat}, a feed-forward framework that reframes occlusion
        handling as Occlusion-Aware Gaussian Densification, targeting missing scene
        geometry rather than missing pixels by predicting new Gaussian primitives for
        occluded regions.
  \item We design two modules realizing this idea: Gaussian Occlusion Mapping, which identifies regions lacking Gaussian support, and Gaussian Occlusion Completion, which generates and integrates new primitives for them.
  \item Extensive experiments on nuScenes show that OccSplat improves on the compared feed-forward methods, with particularly large gains in occluded regions.
\end{itemize}

\section{Related Work}
\label{sec:Related Work}



\textbf{Feed-forward 3D Gaussian Splatting.} To eliminate the expensive per-scene optimization of conventional 3DGS, numerous feed-forward approaches have been proposed to directly predict Gaussian representations from sparse input images. Early methods such as PixelSplat \cite{charatan2024pixelsplat} and Splatter Image \cite{szymanowicz2024splatter} generate Gaussians from image features, while MVSplat \cite{chen2024mvsplat} and DepthSplat \cite{xu2025depthsplat} improve geometric reconstruction through explicit cross-view matching and geometry-aware representations. Subsequent methods~\cite{ye2025no, jiang2025anysplat, zhang2025flare, ye2025yonosplat}, further relax pose requirements and enable reconstruction in pose-free or fully uncalibrated settings. Despite their impressive efficiency and generalization ability, these methods often struggle in sparse-view scenarios with severe occlusions and limited view overlap, resulting in incomplete geometry and inconsistent scene representations.

\textbf{Driving Scene Reconstruction.} NeRFs~\cite{mildenhall2021nerf} and 3DGS~\cite{3dgs} have been adapted to driving scene reconstruction~\cite{zhou2024drivinggaussian, yan2024street, huang2024textit, khan2025autosplat}. However, most of them focus on high-quality reconstruction of individual scenes, relying on offline per-scene optimization, which limits their applicability to online driving systems. Their reconstruction quality can also degrade under severe occlusions and sparse-view observations. A few recent feed-forward and pose-free approaches, such as \cite{ye2025no, smart2024splatt3r, jiang2025anysplat, zhang2025flare}, explore online reconstruction for driving scenes, but they mainly focus on general scene reconstruction and cross-scene generalization, while lacking explicit mechanisms for handling severe occlusions and sparse-view ambiguities. DistillNeRF \cite{wang2024distillnerf} proposes a generalizable framework for driving scenes and achieves reasonable novel view synthesis, but it requires an expensive offline NeRF optimization stage per scene and additional LiDAR supervision, while still struggling in highly occluded regions. In contrast, OccSplat targets feed-forward driving scene reconstruction from RGB inputs alone while improving robustness to occlusions and incomplete observations in sparse-view environments.

\textbf{Occlusion-aware and Structure-aware 3DGS.} Recent efforts have sought to improve structural modeling and visibility reasoning in 3DGS. Scaffold-GS \cite{lu2024scaffold} introduces an anchor-based representation to enhance structural control and reconstruction efficiency, while CityGaussian \cite{liu2024citygaussian} and Octree-GS \cite{ren2024octree} improve scalability through hierarchical scene representations. OccluGaussian \cite{liu2025occlugaussian} further enhances reconstruction consistency by leveraging occlusion-aware scene partitioning based on camera co-visibility. In parallel, methods for recovering missing or occluded regions~\cite{dahaghin2025splatfill, huang20253d, zhou2025high, shi2025imfine} employ depth-guided geometry reasoning, multi-view consistency constraints, and iterative refinement to recover incomplete scene content. However, these methods still rely on scene-specific optimization and therefore cannot operate in a fully feed-forward manner.

\section{Method}
\label{sec:method}

\subsection{Overview}
\label{sec:overview}
\begin{figure*}[t!]
  \centering
  \includegraphics[width=\linewidth]{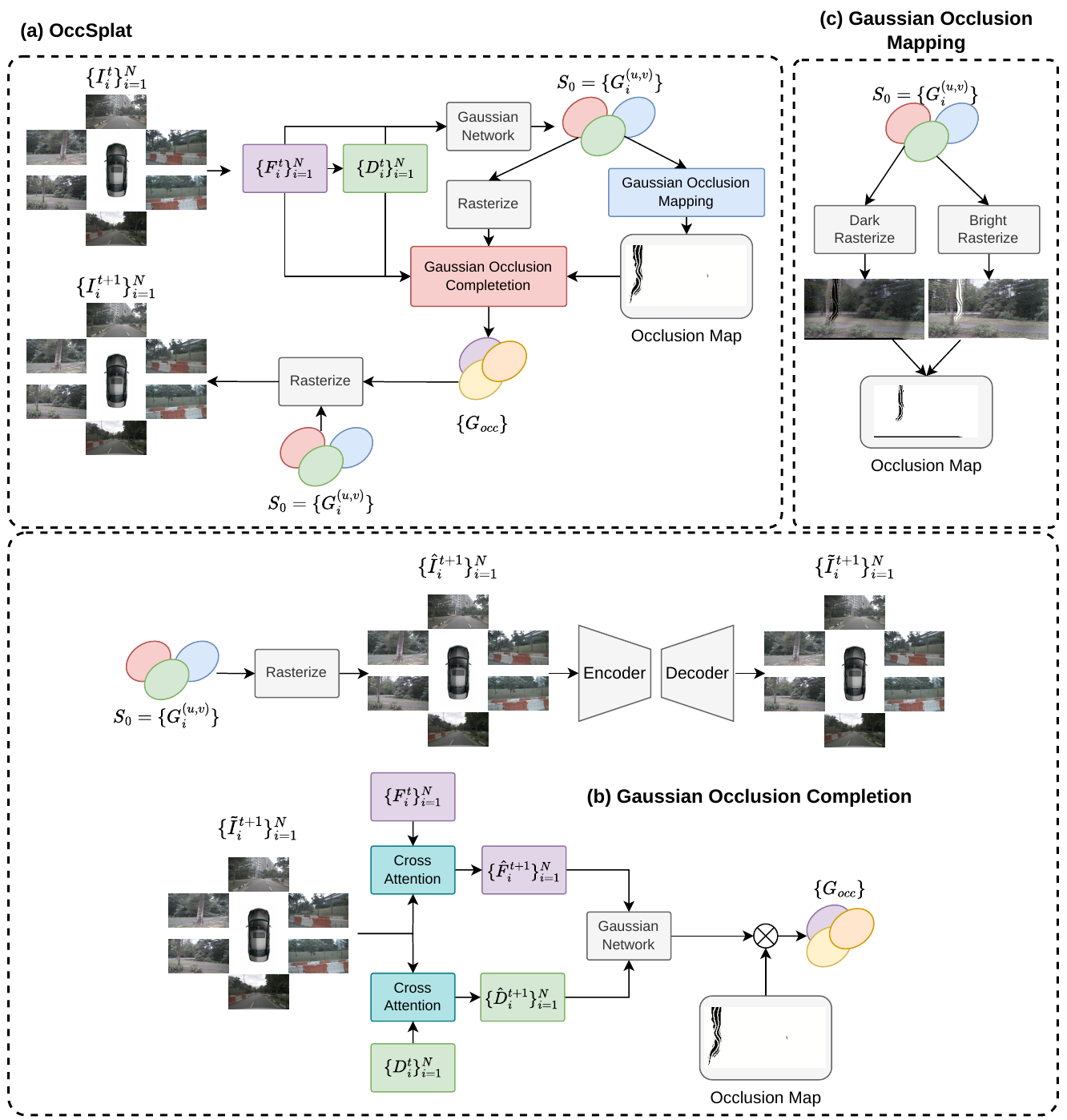}
  \caption{%
  \textbf{Overall architecture of OccSplat.}
  \textbf{(a)} From surround-view images $\{I_i^t\}$, a backbone predicts the pixel-aligned
  Gaussian set $S_0=\{G_i^{(u,v)}\}$, after which Gaussian Occlusion Mapping \textbf{(c)}
  locates unsupported regions by rasterizing $S_0$ against dark and bright backgrounds and
  using their responses to identify unsupported regions.
  \textbf{(b)} Guided by this map, Gaussian Occlusion Completion rasterizes $S_0$ to
  the future view, inpaints it with an encoder-decoder, and lifts the result back into new
  primitives $\{G_{occ}\}$ that densify $S_0$ for rendering the future frame
  $\{I_i^{t+1}\}$.
}
\label{fig:architecture}
\end{figure*}

Given the $N$ surround-view images $\{I_i^t\}_{i=1}^{N}$ of the current frame $t$,
OccSplat aims to synthesize the surround views of the future frame $t{+}1$ while
completing the scene content that the ego-vehicle's forward motion will disocclude. As
shown in Fig.~\ref{fig:architecture}, our pipeline builds on a feed-forward surround-view
backbone and augments it with two occlusion-handling modules that operate directly on the
Gaussian representation.

\textbf{Base Gaussian reconstruction.}
We first reconstruct a base 3D Gaussian representation following
DrivingForward~\cite{tian2025drivingforward}. An encoder extracts multi-scale image
features $\{F_i^t\}_{i=1}^{N}$, which are fused across views through the Volumetric Fusion
Network (VFNet) \cite{kim2022self} to estimate dense depth maps $\{D_i^t\}_{i=1}^{N}$. Each predicted depth
is back-projected into 3D to initialize a Gaussian center,
\begin{equation}
\mu_i^{(u,v)}=\operatorname{Unproject}\!\left(D_i^t(u,v)\right),
\end{equation}
where $\mu_i^{(u,v)}\in\mathbb{R}^3$. The remaining attributes, including opacity
$\alpha$, anisotropic scale $s$, rotation quaternion $q$, and spherical harmonics
coefficients $\mathrm{sh}$, are predicted by a lightweight Gaussian head,
\begin{equation}
\{\alpha,s,q,\mathrm{sh}\}_i^{(u,v)}=\mathcal{G}_{\theta}\!\left([F_i^t,D_i^t]\right).
\end{equation}
Aggregating the primitives from all views yields the pixel-aligned Gaussian set
\begin{equation}
S_0=\{G_i^{(u,v)}\},
\end{equation}
which is rendered through differentiable rasterization.

Because $S_0$ is generated only from pixels observed at $t$, it provides no support for
surfaces occluded in the current frame. As the viewpoint advances, these hidden regions
become visible, yet the renderer lacks the primitives needed to synthesize them, producing
structural holes, stretched surfaces, and appearance inconsistencies. OccSplat closes this
gap in two steps that operate directly on $S_0$. Gaussian Occlusion Mapping
(Sec.~\ref{sec:mapping}) locates the regions that lack Gaussian support, and Gaussian
Occlusion Completion (Sec.~\ref{sec:completion}) generates new primitives $\{G_{occ}\}$ for
those regions and merges them into $S_0$, so that the future frame is rendered from a
densified, more complete representation.

\subsection{Gaussian Occlusion Mapping}
\label{sec:mapping}

The first step is to identify occluded regions that lack Gaussian support. Rather than introducing a separate segmentation network, we exploit the rendering process itself to detect pixels with insufficient Gaussian coverage or accumulated opacity, constructing a spatial occlusion map directly from the Gaussian density.

Specifically, we render the base representation $S_0$ to the target view $\pi_{t+1}$ twice under complementary background colors, producing two sets of surround-view renderings,
\begin{align}
\{I_{i,\text{dark}}^{t+1}\}_{i=1}^{N}
  &= \mathrm{Render}(S_0,\pi_{t+1},\mathrm{bg}=[0,0,0]),\\
\{I_{i,\text{bright}}^{t+1}\}_{i=1}^{N}
  &= \mathrm{Render}(S_0,\pi_{t+1},\mathrm{bg}=[1,1,1]).
\end{align}

Pixels supported by sufficiently dense, opaque Gaussians (high accumulated opacity) render almost identically under both backgrounds. In contrast, pixels with insufficient Gaussian support—either completely unpopulated or covered only by low-opacity primitives—are dominated by the background colors and thus differ strongly between the two renders. Regions with low accumulated coverage are identified as lacking geometric support:

\begin{equation}
\{M_{i,\text{occ}}\}_{i=1}^{N}
=\mathbbm{1}[\{I_{i,\text{bright}}^{t+1}\}_{i=1}^{N}>\epsilon]
\land
\mathbbm{1}[\{I_{i,\text{dark}}^{t+1}\}_{i=1}^{N}<(1-\epsilon)].
\end{equation}
This rendering-based strategy produces an occlusion map without additional
learnable parameters and restricts Gaussian generation to unsupported regions.

\subsection{Gaussian Occlusion Completion}
\label{sec:completion}

Given the occlusion masks $\{M_{i,\text{occ}}\}$, the next step is to generate new Gaussian
primitives that fill the unsupported regions. Predicting these primitives directly from
the raw future-view rendering is unreliable, because disocclusion artifacts corrupt both
local appearance and structural continuity and introduce noisy cues for the subsequent
geometry inference. We therefore first restore the rendering, then decouple geometry and
appearance completion into two linear attention \cite{katharopoulos2020transformers} branches, and finally reuse the
pretrained Gaussian head to produce the new primitives.

\textbf{Pixel Refiner.}
We first rasterize $S_0$ to the future viewpoint to obtain a preliminary render
$\hat I_i^{t+1}$ that exhibits the characteristic disocclusion holes. A lightweight
encoder-decoder (U-Net) then restores plausible image structure in these regions while
preserving already well-reconstructed areas, predicting a residual correction to the
rendered view,
\begin{equation}
\tilde{I}_i^{t+1}=\mathrm{clamp}\!\left(\hat{I}_i^{t+1}+\mathcal{R}_{\phi}(\hat{I}_i^{t+1}),0,1\right),
\end{equation}
where $\hat{I}_i^{t+1}$ and $\tilde{I}_i^{t+1}$ are the rendered and refined future-view
images and $\mathcal{R}_{\phi}$ is the residual network. The residual formulation is designed to restore missing structure while preserving
regions already reconstructed by $S_0$.
The refined image is not the final output, but rather a cleaner intermediate cue that
enables more reliable depth and feature inference for the completion below.

\textbf{Decoupled geometry and appearance inference.}
Recovering a missing primitive requires estimating both its 3D location and its visual
attributes, which rely on different cues: geometry completion depends on depth consistency
across views, whereas appearance completion aggregates semantic and texture information
from the source frame. To reduce potential interference between these tasks, we use separate branches. The refined future view
$\tilde{I}_i^{t+1}$ serves as the query, while the source-frame depth $\{D_i^t\}$ and
multi-scale features $\{F_i^t\}$ provide the key--value representations. We employ a Depth
Attention module $\mathcal{A}_d$ for geometry and a Feature Attention module
$\mathcal{A}_f$ for appearance,
\begin{align}
\hat D_i^{t+1} &= \mathcal A_d\!\left(\tilde I_i^{t+1},\,D_i^t,\,D_i^t\right),\\
\hat F_i^{t+1} &= \mathcal A_f\!\left(\tilde I_i^{t+1},\,F_i^t,\,F_i^t\right).
\end{align}
Each branch first applies self-attention to aggregate context across source views, then
cross-attention to transfer it to the future-view query. Using linear attention \cite{katharopoulos2020transformers} reduces
complexity from $\mathcal{O}(n^2)$ to $\mathcal{O}(n)$, enabling efficient long-range
correspondence at surround-view resolution.

\textbf{Gaussian generation and fusion.}
The inferred depth determines the positions of the new primitives. Each pixel inside
the occluded region is back-projected using this depth to initialize a Gaussian center,
\begin{equation}
\mu_{\text{occ}}^{(u,v)}=\operatorname{Unproject}\!\left(\hat D_i^{t+1}\right).
\end{equation}
Rather than training a second attribute predictor, we reuse the Gaussian head
$\mathcal{G}_\theta$ learned during base reconstruction, which already provides a stable
mapping from image features to Gaussian attributes,
\begin{equation}
\{\alpha,s,q,\mathrm{sh}\}_{\text{occ}}^{(u,v)}=\mathcal G_\theta\!\left([\hat F_i^{t+1},\hat D_i^{t+1}]\right).
\end{equation}
Reusing the pretrained head lets the completion branches focus solely on recovering the
missing geometry and appearance instead of relearning attribute estimation. The resulting
primitives, masked by $\{M_{i,\text{occ}}\}$, form the occlusion Gaussians $\{G_{occ}\}$ and
are merged with the base representation to render the future frame,
\begin{align}
S_1 &= S_0 \cup \{G_{occ}\},\\
\{I_i^{t+1}\}_{i=1}^{N} &= \mathrm{Render}(S_1),
\end{align}
providing denser geometric coverage in previously unseen regions for improved novel-view synthesis.

\subsection{Training Objective}
\label{sec:loss}

Optimizing the whole framework at once is unstable, since the completion branches depend
on reliable geometry and appearance from the base network. During early training,
inaccurate depth and unstable Gaussians provide noisy supervision. We therefore adopt a
two-stage strategy that first learns a stable Gaussian representation and then introduces
occlusion-aware densification.

\textbf{Stage I: base reconstruction.}
We optimize the depth network, pose network, Gaussian head, and Pixel Refiner, keeping the completion branches frozen. The objective follows DrivingForward~\cite{tian2025drivingforward} with an additional refinement loss.
\begin{equation}
\begin{split}
\mathcal L_{\text{stage1}}
={}&\mathcal L_{\text{reproj}}
+\lambda_{\text{sm}}\mathcal L_{\text{smooth}}
+\lambda_{\text{sp}}\mathcal L_{\text{spatio}}
+\lambda_{\text{st}}\mathcal L_{\text{st}}\\
&+\lambda_{\text{gs}}\mathcal L_{\text{gs}}
+\lambda_{\text{ref}}\mathcal L_{\text{refine}},
\end{split}
\end{equation}

\begin{equation}
\mathcal{L}_{\text{refine}}
=\lambda_{\text{MSE}}\|\tilde{I}_i^{t+1}-I_i^{t+1}\|_2^2
+\lambda_{\text{LPIPS}}\mathcal{L}_{\text{LPIPS}}(\tilde{I}_i^{t+1},I_i^{t+1}).
\end{equation}
The MSE term enforces pixel-level fidelity while the perceptual term encourages structurally coherent restoration in disoccluded regions.

\textbf{Stage II: occlusion-aware densification.}
Once the base reconstruction converges, all Stage-I modules are frozen and only the Depth Attention and Feature Attention branches are optimized. The Gaussian head is reused without fine-tuning to preserve the learned feature-to-Gaussian mapping. The objective augments the Stage-I loss with supervision on the inferred future-view depth and features,
\begin{equation}
\mathcal L_{\text{stage2}}
=\mathcal L_{\text{stage1}}
+\lambda_{\text{nd}}\mathcal L_{\text{novel-depth}}
+\lambda_{\text{nf}}\mathcal L_{\text{novel-feature}}.
\end{equation}

The inferred depth $\hat D_i^{t+1}$ is supervised with target-view pseudo-depth predicted by Depth Anything V2~\cite{yang2024depthanythingv2}, and the inferred features $\hat F_i^{t+1}$ are supervised with detached encoder features of the corresponding ground-truth image.

\section{Experiments}
\subsection{Experimental Setup}
\textbf{Datasets.}
We evaluate on the nuScenes dataset~\cite{caesar2020nuscenes}, comprising 1,000 driving scenes with six surround-view cameras capturing a $360^{\circ}$ view with ${\sim}10\%$ overlap between adjacent views. Following the official split, we use 700 scenes for training and 150 for validation at $114 \times 228$ resolution.

\textbf{Evaluation Metrics.}
We evaluate novel-view synthesis quality at time step $t+1$ given observations at $t$, reporting standard PSNR ($\uparrow$), SSIM~\cite{wang2004image} ($\uparrow$), and LPIPS~\cite{zhang2018unreasonable} ($\downarrow$).

\textbf{Implementation Details.} Training has two stages. Stage I jointly optimizes the base network and Pixel Refiner with DrivingForward losses and $\mathcal{L}_{\text{refine}}$ ($\lambda_{\text{ref}}=0.05, \lambda_{\text{MSE}}=1.0, \lambda_{\text{LPIPS}}=0.05$). Stage II freezes Stage I weights and trains $\mathcal{A}_d$ and $\mathcal{A}_f$ under novel-view depth ($\lambda_{\text{nd}}=1.0$) and feature supervision ($\lambda_{\text{nf}}=0.5$). We set $\epsilon=0.7$, depth range $[1.5,80.0]\,\text{m}$, learning rate $1\times10^{-4}$, and batch size 1. Both stages run for 10 epochs on an NVIDIA RTX 4090.

\subsection{Quantitative Results on nuScenes}

We quantitatively compare our approach with prior novel view synthesis methods on the nuScenes validation set, as summarized in Table \ref{feed_forward}. Following DrivingForward \cite{tian2025drivingforward}, all methods are evaluated at a resolution of $114 \times 228$. OccSplat achieves the best overall performance. Compared with DrivingForward, our approach improves PSNR by \textbf{1.74} dB while reducing LPIPS from 0.194 to \textbf{0.182}, indicating better reconstruction fidelity and perceptual quality.

The comparison includes NeRF- and occupancy-based methods such as SelfOcc \cite{huang2024selfocc}, UniPad \cite{yang2024unipad}, EmerNeRF \cite{yang2024emernerf}, and DistillNeRF \cite{wang2024distillnerf}, as well as feed-forward 3DGS methods such as MVSplat \cite{chen2024mvsplat} and DrivingForward \cite{tian2025drivingforward}. These results highlight the effectiveness of occlusion-aware Gaussian densification in recovering missing scene content and improving geometric completeness, leading to more accurate novel-view rendering in challenging surround-view driving scenarios.

\begin{table}[t]
\centering
\footnotesize
\caption{Quantitative comparison of our method (OccSplat) against prior reconstruction methods on the nuScenes validation set. \textbf{Best} in bold, \underline{second best} underlined.}
\label{feed_forward}
\setlength{\tabcolsep}{12pt}
\begin{tabular}{l|c|ccc}
\toprule
Method & Venue & PSNR$\uparrow$ & SSIM$\uparrow$ & LPIPS$\downarrow$ \\
\midrule
SelfOcc \cite{huang2024selfocc} & CVPR 2024 & 18.22 & 0.464 & -- \\
UniPad \cite{yang2024unipad} & CVPR 2024 & 16.45 & 0.375 & -- \\
EmerNeRF \cite{yang2024emernerf} & ICLR 2024 & 20.95 & 0.585 & -- \\
DistillNeRF \cite{wang2024distillnerf} & NeurIPS 2024 & 20.78 & 0.590 & -- \\
MVSplat \cite{chen2024mvsplat} & ECCV 2024 & 20.23 & 0.479 & 0.366 \\
DrivingForward \cite{tian2025drivingforward} & AAAI 2025 & \underline{21.76} & \underline{0.767} & \underline{0.194} \\
OccSplat (Ours) & -- & \textbf{23.50} & \textbf{0.767} & \textbf{0.182} \\
\bottomrule
\end{tabular}
\end{table}

\subsection{Ablation Studies}
To validate the effectiveness of each component, we perform ablation experiments on the
nuScenes validation set, progressively adding Gaussian Occlusion Completion, the Pixel
Refiner, and Gaussian Occlusion Mapping on top of the base reconstruction. Quantitative
results are reported in Table~\ref{tab:ablation}.
\begin{table}[t]
\centering
\footnotesize
\caption{Ablation study of the proposed components on the nuScenes validation set. We
report PSNR ($\uparrow$), SSIM ($\uparrow$), and LPIPS ($\downarrow$). ``Densification''
denotes Gaussian Occlusion Completion,
``Refiner'' the Pixel Refiner, and ``Occ.\ Mapping'' the dual-background Gaussian
Occlusion Mapping.}
\label{tab:ablation}
\begin{tabular*}{\linewidth}{@{\extracolsep{\fill}}c|ccc|ccc}
\toprule
Config. & Densification & Refiner & Occ.\ Mapping & PSNR$\uparrow$ & SSIM$\uparrow$ & LPIPS$\downarrow$ \\
\midrule
(a) & & & & 21.76 & \textbf{0.767} & 0.194 \\
(b) & \checkmark & & & 19.92 & 0.747 & 0.182 \\
(c) & \checkmark & \checkmark & & 20.71 & 0.753 & \textbf{0.175} \\
(d) & \checkmark & & \checkmark & 22.27 & 0.765 & 0.187 \\
(e) & \checkmark & \checkmark & \checkmark & \textbf{23.50} & \textbf{0.767} & 0.182 \\
\bottomrule
\end{tabular*}
\end{table}

\textbf{Gaussian Densification.}
Comparing (a) and (b), applying Gaussian Occlusion Completion across the entire image lowers
PSNR from 21.76\,dB to 19.92\,dB and SSIM from 0.767 to 0.747, while slightly improving LPIPS
from 0.194 to 0.182. Predicting new primitives from the raw, artifact-corrupted future-view
renders and inserting them everywhere adds redundant Gaussians in already reconstructed areas,
degrading pixel-wise fidelity even though the added geometry recovers some perceptual detail.
This result motivates using reliable completion cues and restricting densification
to unsupported regions.

\textbf{Pixel Refiner.}
Adding the Pixel Refiner in (c) improves all three metrics over (b), raising PSNR to
20.71\,dB and SSIM to 0.753 while achieving the best LPIPS of 0.175. By restoring plausible
appearance in the disoccluded areas before completion, the Refiner provides a cleaner
intermediate representation, so the depth and feature attention modules infer more accurate
geometry for the missing regions. PSNR and SSIM nonetheless remain below those of the base reconstruction,
since densification is still applied over the whole image without localization.

\textbf{Occlusion Mapping.}
Gaussian Occlusion Mapping supplies the missing localization. Even without the Refiner,
restricting densification to disoccluded regions in (d) already lifts PSNR to 22.27\,dB, above
the base reconstruction. Combining all components in (e) yields the best overall result,
raising PSNR to 23.50\,dB and restoring SSIM to 0.767 while retaining strong perceptual
quality. By identifying disoccluded regions through dual-background rendering, the mapping
confines completion to where the representation truly lacks support, preserving the original
reconstruction elsewhere and supporting the complementary roles of densification, the Refiner, and the mapping.

\textbf{Effect of the Occlusion Threshold $\epsilon$.}
The occlusion threshold $\epsilon$ in Eq.~(6) controls the sensitivity of Gaussian Occlusion Mapping when identifying unsupported regions. A larger threshold produces a more conservative occlusion mask, causing fewer pixels to be classified as unsupported and therefore limiting Gaussian generation. As a result, many partially disoccluded areas remain unrecovered, leading to lower reconstruction quality. This behavior is reflected by the inferior performance at $\epsilon=0.99$ (22.91\,dB PSNR and 0.187 LPIPS).

Reducing the threshold allows more missing regions to be recovered, improving reconstruction quality and reaching the best performance at $\epsilon=0.7$ (23.50\,dB PSNR and 0.182 LPIPS). However, further decreasing $\epsilon$ causes the occlusion mask to become overly permissive, generating redundant Gaussian primitives in regions that are already well reconstructed. This introduces unnecessary geometric redundancy and degrades rendering quality, as evidenced by the performance drop at $\epsilon=0.6$ and $0.5$. These results indicate that $\epsilon=0.7$ provides the best trade-off between recovering genuinely missing content and avoiding excessive Gaussian generation.

\begin{table}[t]
\centering
\caption{Effect of the occlusion threshold $\epsilon$ in Gaussian Occlusion Mapping.}
\label{tab:epsilon}

\renewcommand{\arraystretch}{1.15}
\setlength{\tabcolsep}{10pt}

\begin{tabular}{c|ccc}
\toprule
$\epsilon$ & PSNR$\uparrow$ & SSIM$\uparrow$ & LPIPS$\downarrow$ \\
\midrule
0.99 & 22.91 & 0.764 & 0.187 \\
0.8 & 23.41 & \textbf{0.767} & 0.183 \\
\textbf{0.7} & \textbf{23.50} & \textbf{0.767} & \textbf{0.182} \\
0.6 & 23.30 & 0.757 & 0.192 \\
0.5 & 22.55 & 0.741 & 0.204 \\
\bottomrule
\end{tabular}
\end{table}

\textbf{Evaluation on Disoccluded Regions.}
To isolate the effect of our method, we additionally evaluate only disoccluded regions using identical occlusion masks for all methods, as shown in \cref{tab:disoccluded_metric}. DrivingForward achieves only 11.498\,dB PSNR due to the lack of Gaussian primitives in disoccluded areas. While MVSplat achieves a comparable PSNR of 18.606\,dB (vs.\ 18.602\,dB for OccSplat), its erroneous depth estimation causes severe structural distortion and scattered artifacts, leading to a drastic drop in SSIM (0.509). In contrast, OccSplat preserves structural fidelity while completing missing geometry, achieving the highest SSIM (\textbf{0.763}) and lowest LPIPS (\textbf{0.465}) among the compared methods.

\begin{table}[t]
\centering
\footnotesize
\caption{Performance on disoccluded regions.}
\label{tab:disoccluded_metric}
\setlength{\tabcolsep}{12pt}
\begin{tabular}{lccc}
\toprule
Method & PSNR $\uparrow$ & SSIM $\uparrow$ & LPIPS$ \downarrow$ \\
\midrule
DrivingForward \cite{tian2025drivingforward} & 11.498 & 0.759 & 1.178 \\
MVSplat \cite{chen2024mvsplat} & \textbf{18.606} & 0.509 & 0.620 \\
OccSplat (Ours) & 18.602 & \textbf{0.763} & \textbf{0.465} \\
\bottomrule
\end{tabular}
\end{table}

\begin{table}[t]
\centering
\caption{Computational costs. Time denotes per-scene optimization for 3DGS and inference for the feed-forward methods.}
\label{tab:efficiency}
\setlength{\tabcolsep}{8pt}
\begin{tabular}{lcccc}
\toprule
Method & FPS $\uparrow$ & Time $\downarrow$ & VRAM (GB) $\downarrow$ & Params (M) $\downarrow$ \\
\midrule
3DGS & -- & $\sim$9 min & -- & -- \\
DrivingForward & \textbf{1.74} & \textbf{574.7 ms} & \textbf{2.96} & \textbf{61.88} \\
OccSplat & 1.01 & 992.9 ms & 3.10 & 63.16 \\
\bottomrule
\end{tabular}
\end{table}

\textbf{Efficiency Comparison.}
OccSplat retains the feed-forward advantage over per-scene 3DGS optimization. As reported in Table~\ref{tab:efficiency}, compared with DrivingForward, OccSplat incurs modest additional costs (+0.14GB VRAM and +1.28M parameters) while achieving substantial reconstruction quality gains at approximately 1.01 FPS. Unlike conventional per-scene 3DGS methods, which require iterative optimization for each scene, OccSplat performs feed-forward reconstruction without scene-specific optimization.



\subsection{Qualitative Results}

\begin{figure}[t] 
\centering 

\includegraphics[width=\linewidth]{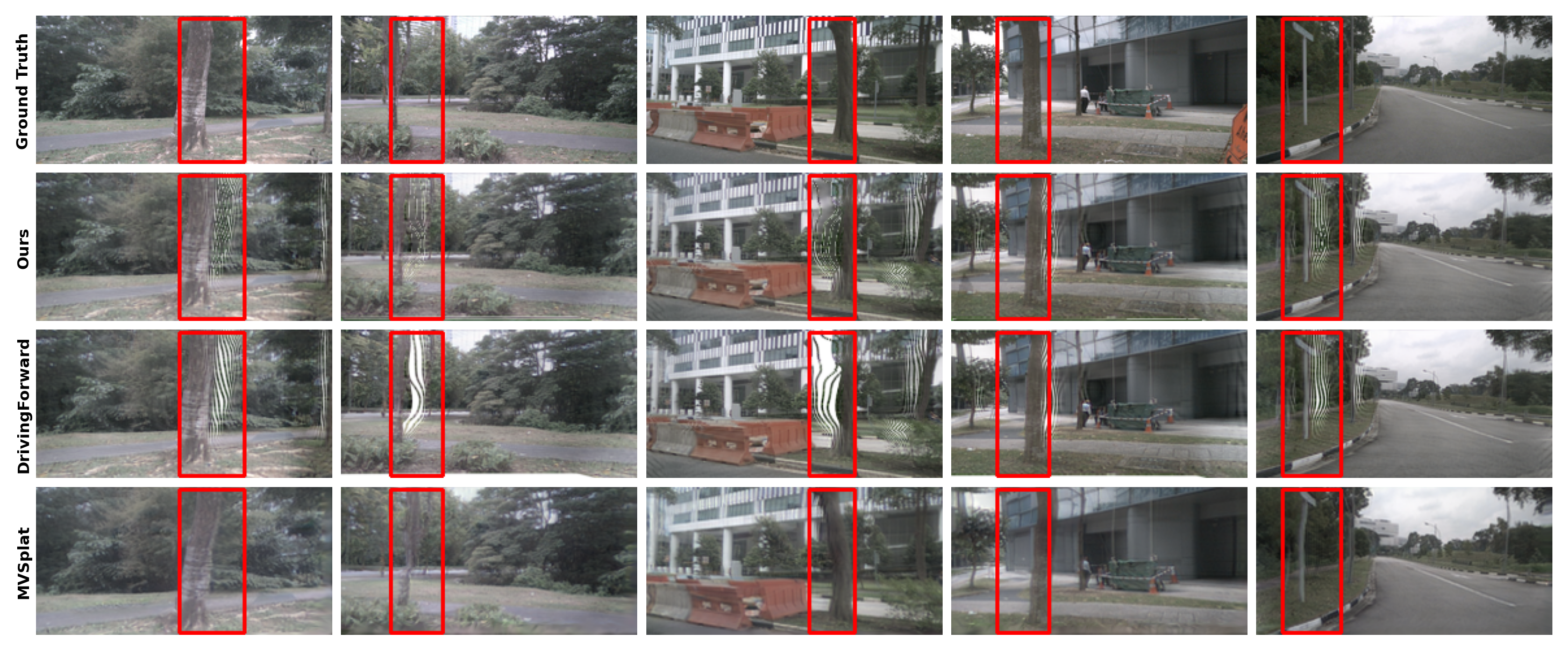} 
\caption{Qualitative comparison on the nuScenes validation set. From top to bottom: Ground Truth, OccSplat (ours), DrivingForward, and MVSplat. The highlighted regions show that OccSplat recovers newly disoccluded content while preserving existing scene structures, whereas DrivingForward leaves holes due to missing Gaussian support and MVSplat exhibits geometric distortions caused by inaccurate depth estimation.}

\label{fig:qualitative} 
\end{figure}

Figure~\ref{fig:qualitative} presents qualitative comparisons among MVSplat, DrivingForward, and OccSplat on representative scenes from the nuScenes validation set. MVSplat suffers from inaccurate geometry caused by erroneous depth estimation at the source frame, leading to distorted structures and inconsistent renderings in the future view. DrivingForward reconstructs the visible scene more accurately, but when the viewpoint moves to frame $t+1$, newly disoccluded regions lack Gaussian support, resulting in holes and stretched artifacts around tree trunks, building boundaries, and roadside structures.

OccSplat explicitly addresses this failure mode by generating Gaussian primitives only in unsupported disoccluded regions while preserving the correctly reconstructed geometry. As highlighted in the red boxes, OccSplat recovers missing content and produces renderings that are visually closer to the ground truth. These observations are consistent with the quantitative improvements reported in Table~\ref{feed_forward}, demonstrating the effectiveness of the proposed occlusion-aware Gaussian densification.

\begin{figure}[H] \centering \includegraphics[width=\linewidth]{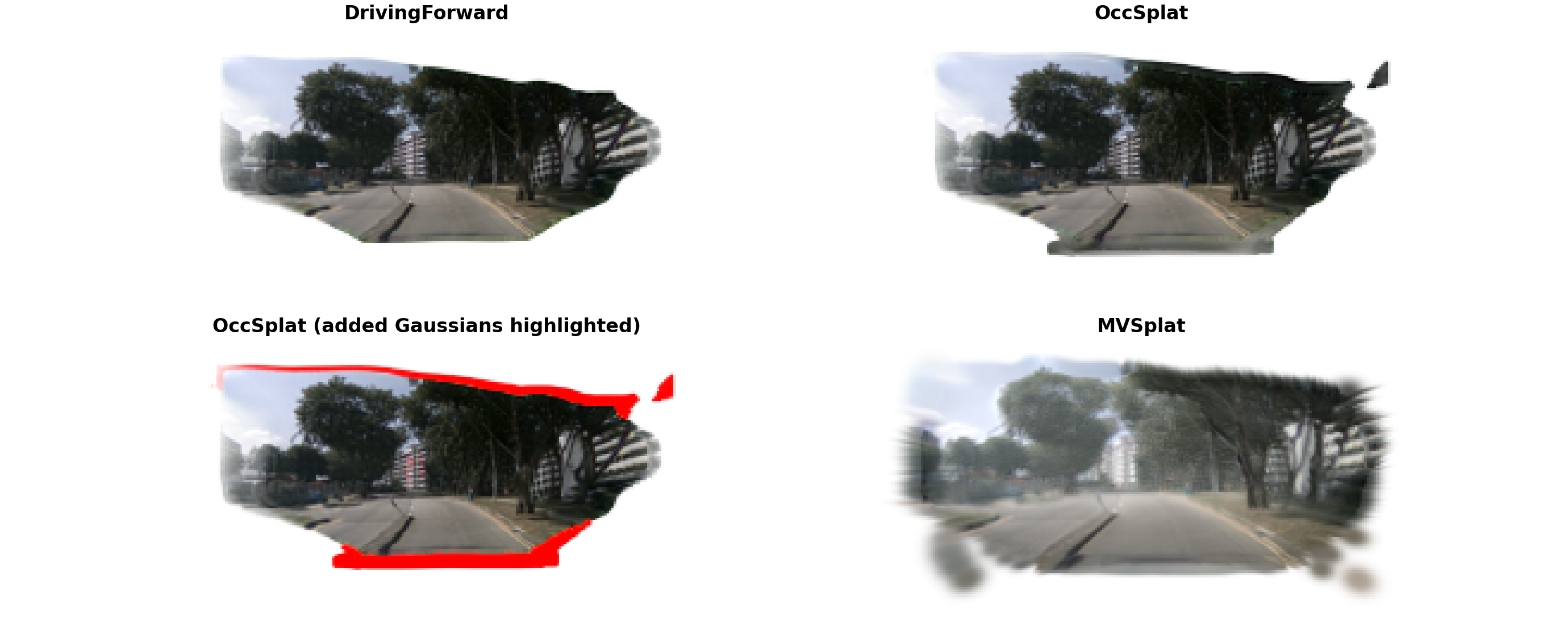} \caption{Qualitative comparison of reconstructed 3D Gaussian scenes. Top row: DrivingForward (left) and OccSplat (right). Bottom row: OccSplat with newly added Gaussian primitives highlighted in red (left) and MVSplat (right). The added primitives are distributed along the upper scene boundary, the road foreground, and other regions with incomplete coverage.} \label{fig:qualitative_3d} \end{figure}

\textbf{3D Gaussian Scene Comparison.} Figure~\ref{fig:qualitative_3d} compares the reconstructed 3D Gaussian scenes. DrivingForward preserves the main scene structures but exhibits incomplete coverage in occluded regions. In contrast, MVSplat produces less coherent structures, with noticeable blurring and scattered Gaussian artifacts around the reconstructed scene. OccSplat extends the scene coverage while maintaining the overall structural consistency of the base reconstruction. The bottom-left panel highlights the added Gaussian primitives in red, showing their distribution along the upper scene boundary, the road foreground, and other regions with incomplete coverage. These results qualitatively demonstrate how Gaussian Occlusion Completion supplements the existing Gaussian representation to improve scene coverage.

\section{Conclusion}

We presented \textbf{OccSplat}, an occlusion-aware feed-forward framework that addresses disocclusion in single-frame novel-view synthesis for driving scenes. By identifying disoccluded regions and generating complementary primitives to densify the 3D Gaussian representation, OccSplat improves scene coverage while retaining feed-forward inference. Experiments on nuScenes demonstrate improved reconstruction quality relative to the compared methods, particularly in newly exposed areas. Furthermore, the completion strategy offers a potential extension to other feed-forward 3DGS models.

\textbf{Limitations and Future Work.} Operating in image space, the Pixel Refiner cannot explicitly enforce 3D geometric consistency across viewpoints, leading to potential texture over-smoothing under severe disocclusions or large ego-motion that reveals previously unobserved content. Future work will integrate explicit 3D geometric and semantic priors, as well as temporal information from consecutive frames, to improve structural consistency and robustness in dynamic driving environments.

\section*{Acknowledgement}
We acknowledge Ho Chi Minh City University of Technology (HCMUT), VNU-HCM for supporting this study.

\bibliographystyle{splncs04}
\bibliography{main}
\end{document}